\documentclass[letterpaper]{article} 
\usepackage[preprint]{aaai2027}  

\usepackage[hyphens]{url}  
\usepackage{graphicx} 
\usepackage{natbib}  
\usepackage{caption} 
\usepackage{algorithm}
\usepackage{algorithmic}
\usepackage{amsmath}
\usepackage{amssymb}
\usepackage{amsthm}
\usepackage{microtype}

\usepackage{latexsym}

\usepackage{minitoc}
\usepackage[many]{tcolorbox}
\usepackage{longtable}
\usepackage{enumitem}

\usepackage[T1]{fontenc}

\usepackage[utf8]{inputenc}

\usepackage{inconsolata}

\usepackage{xspace}

\usepackage{booktabs}   
\usepackage{multirow}   
\usepackage{array}      
\usepackage{arydshln}   

\usepackage{colortbl}
\usepackage{pgf}
\usepackage{tabularray}
\usepackage{tabularx}
\usepackage{adjustbox}
\usepackage{newfloat}
\usepackage{listings}

\newcommand{\pila}{PILA\xspace}

\DeclareCaptionStyle{ruled}{labelfont=normalfont,labelsep=colon,strut=off} 
\floatstyle{ruled}
\newfloat{listing}{tb}{lst}{}
\floatname{listing}{Listing}

\title{\pila: Plug-and-Play Insertion for LLM-native Advertising}
\author{
    Zhaowei Zhang\textsuperscript{\rm 1}\equalcontrib \ Yuhan Fu\textsuperscript{\rm 2}\equalcontrib \ Yihang Zhang\textsuperscript{\rm 2} \ Xiaohan Liu\textsuperscript{\rm 3} \ Ceyao Zhang\textsuperscript{\rm 1} \ Xiaoyuan Zhang\textsuperscript{\rm 1} \ Yipeng Kang\textsuperscript{\rm 4}\corresponding \ Tonghan Wang\textsuperscript{\rm 2 \rm 5}\corresponding \ Yaodong Yang\textsuperscript{\rm 1}
}
\affiliations{
    \textsuperscript{\rm 1}Institute for AI, Peking University \ \textsuperscript{\rm 2}College of AI, Tsinghua University \
    \textsuperscript{\rm 3}University of Michigan \
    \textsuperscript{\rm 4}State Key Laboratory of General
Artificial Intelligence, BIGAI \
    \textsuperscript{\rm 5}Shanghai Qi Zhi Institute \

}

\begin{document}

\maketitle

\begin{abstract}
Monetizing large language models (LLMs) by naturally integrating sponsored content into their responses, known as LLM-native advertising, has recently emerged as a critical problem. However, existing solutions entangle advertising with content generation inside a single model, which is incompatible with modern API-only or workflow-based agentic LLM applications and inevitably compromises the original response quality. To address this, we propose \pila, which reformulates ad insertion as a conditional response rewriting problem and decouples it from the upstream service as a lightweight sidecar module. \pila is model-agnostic and can be seamlessly integrated with existing LLM services without modifying the base model or workflow. It also provides a controllable trade-off between user-side naturalness and ad-side exposure, offering a practical interface for downstream pricing and deployment. Experiments across diverse upstream models show that \pila consistently improves ad effectiveness while preserving response quality, highlighting its promise as a practical solution for LLM-native advertising.
\end{abstract}


\section{Introduction}

The rapid advances in large language models (LLMs) have positioned them as key gateways for user traffic across a wide range of applications, including generative search \citep{aggarwal2024geo, wu2025generative}, content generation \citep{achiam2023gpt}, and increasingly widespread deployment of LLM agentic systems \citep{claudecode, codex, openclaw}. 
This shift has given rise to new monetization strategies, most notably the emerging paradigm of LLM-native advertising \citep{zhao2025llm, zhang2026naiad, yun2026llm}, where advertisements are integrated into model-generated responses. 
Ongoing controversy \citep{observer2026openaiads, ndtv2026backlash} over its impact on user experience has made the design of native ad insertion mechanisms that preserve response quality an important research problem across the LLM application ecosystem, including for leading AI companies like OpenAI \citep{openai_ad}. 

Existing approaches to native LLM advertising largely intervene in the generation process itself, by fine-tuning model parameters \citep{zhao2025llm} or exploiting internal model mechanisms \citep{duetting2024mechanism,yun2026llm}, thereby jointly generating the response and the advertisement. This coupled paradigm faces several practical limitations. (1) Frontier LLMs are often closed-source and accessible only through APIs, leaving their parameters and internal states unavailable to downstream developers. (2) Real-world applications are increasingly built from agentic workflows \citep{zhuge2024gptswarm, zhang2025aflow} or harness systems \citep{agent_harness, harness_engineering}, where advertising must operate across heterogeneous components rather than inside a single model. (3) Asking the same model to optimize both response quality and advertising may interfere with the primary objective of serving the user. These limitations point to a common architectural bottleneck: existing methods tightly couple content generation with ad generation. This motivates a decoupled formulation, where ad insertion is formulated as an external, model-agnostic capability that can be applied after, rather than during, response generation. 

\begin{figure*}[!th]
  \centering
  \includegraphics[width=\textwidth]{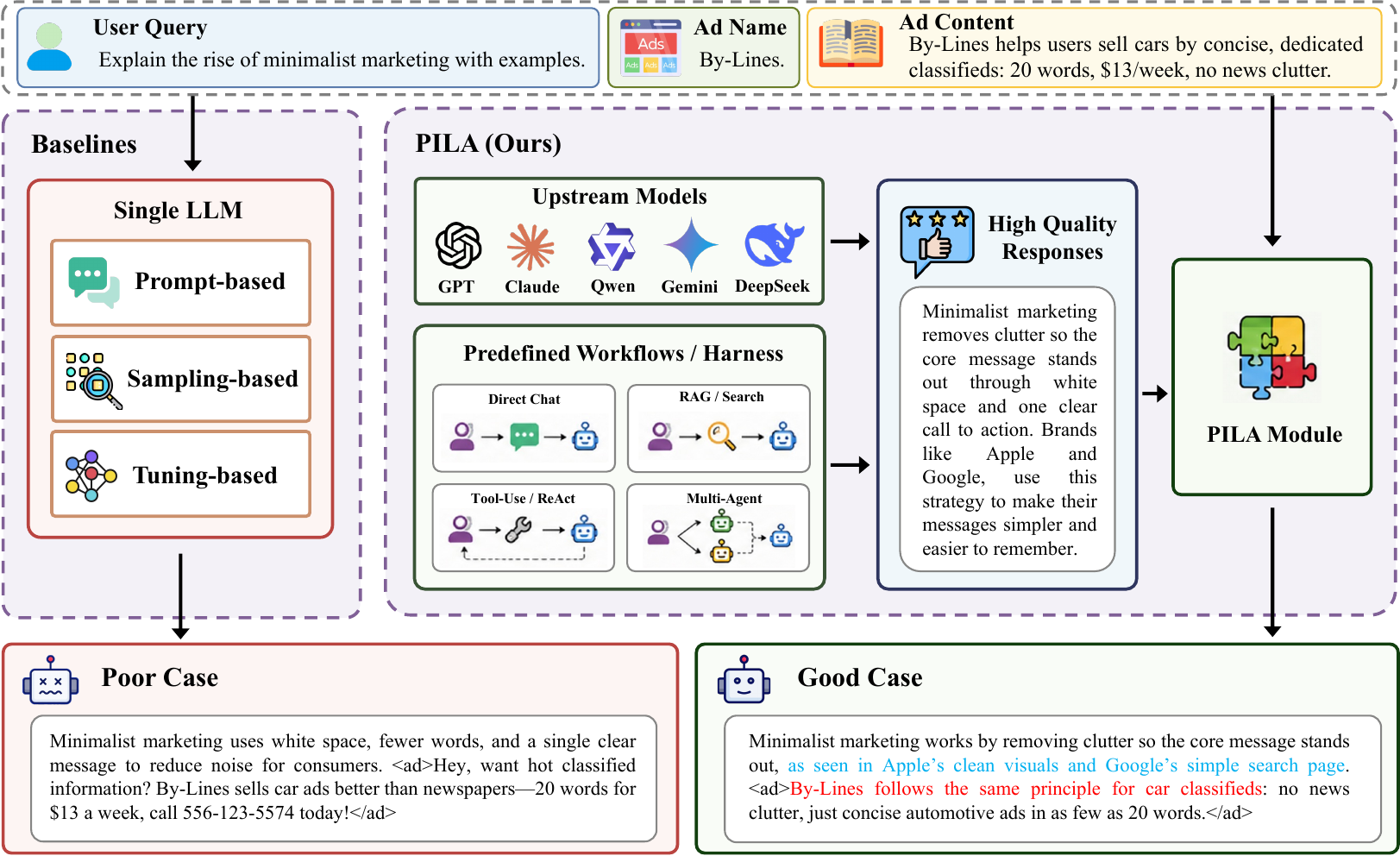}
  \caption{\textbf{An overview of our \pila framework.} Given a user query along with an ad name and ad content, existing baselines either rely on a single LLM with prompting, sampling, or tuning strategies, or follow predefined workflows that often produce poor cases while losing flexibility. In contrast, \pila acts as a lightweight plug-and-play module that adapts diverse upstream models to generate high-quality, naturally integrated responses, yielding better cases in which ad content is seamlessly woven into the user-facing responses. Blue and red highlights in cases indicate that \pila improves quality on both the user and ad side.}
  \label{fig:main}
\end{figure*}

Guided by this principle, we propose \textbf{P}lug-and-play \textbf{I}nsertion for \textbf{L}LM-native \textbf{A}dvertising (\pila), a simple yet highly effective framework that addresses the above challenges at once. 
Inspired by the sidecar pattern in cloud-native architectures \citep{burns2016design}, \pila keeps the upstream systems (LLM models or agentic workflows) frozen and delegates advertising to a lightweight external model.
As shown in Figure \ref{fig:main}, using natural language as the only communication interface, \pila can be seamlessly attached to various upstream systems, intervening only at the final-response stage to determine when, where, and how an ad should be inserted, while preserving the generation ability of the original system.
In this way, we formulate ad insertion as a conditional response rewriting problem. To instantiate \pila, we construct a 25k high-quality corpus on top of the NaiAD infrastructure \citep{zhang2026naiad} through quality-controlled synthesis, self-judging, and diversity augmentation; fine-tune Qwen backbones as the rewriter; and introduce an intensity controller grounded in persuasion knowledge model (PKM), which exposes an ad-intensity factor $\rho$ to balance user-side satisfaction and ad-side exposure.

We conduct extensive experiments in various settings, including plug-and-play enhancement over diverse commercial models, controllable ad-intensity adjustment, and comparisons against single-LLM baselines. Results show that \pila achieves the best performance in terms of both user-side quality and ad-side utility, outperforming prompt-, sampling-, and tuning-based baselines by 34.2\%, 47.3\%, and 7.7\% on average, respectively. As a general plug-and-play module, PILA-4B and PILA-8B further improve seven frontier commercial upstream models by 17.2\% and 18.4\%, respectively, moving them toward a more favorable user-side and ad-side trade-off region. Moreover, the ad-intensity controller offers a controllable deployment-time interface for flexibly navigating this trade-off.

In summary, our contributions are three-fold. 
\textbf{First}, we formulate LLM-native ad insertion as conditional response rewriting and propose \pila, a plug-and-play, model-agnostic sidecar framework that can operate over various upstream LLMs and agentic workflows. \textbf{Second}, we present a practical recipe for building \pila, including a 25k-sample data pipeline and a PKM-grounded contrastive-decoding controller for tunable ad intensity. \textbf{Third}, extensive experiments show that \pila consistently improves ad-side effectiveness while preserving user-side satisfaction, demonstrating its practicality for real-world deployment.


\section{Related Works}

In this section, we provide background on the relevant research. We first introduce the concept of LLM-native advertising, and then review existing work on LLM-based auctions.

\paragraph{LLM-native advertising.}
As LLMs increasingly serve as user-facing information portals, advertising for LLMs is shifting from selling predefined ad slots toward naturally integrating sponsored content into generated responses, which is termed as \emph{LLM-native advertising} \citep{zhao2025llm, zhang2026naiad}. 
Existing studies have explored different ways to incorporate advertisements into LLM responses naturally. 
\citet{xu2026ad} propose a ``hard ad insertion'' approach, which directly inserts ads at fixed positions in the response. 
\citet{hajiaghayi_ad_2024, liu2026incentive} use prompt engineering to instruct LLMs to insert advertisements according to given requirements. 
\citet{soumalias_truthful_2025} propose a sampling-based decoding strategy to search for responses that better optimize advertising revenue. 
However, these works remain far from practical. Recently, \citet{zhao2025llm} propose a learning-based generative auction framework that jointly models ad allocation, pricing, and native response generation. 
Nevertheless, its end-to-end learning-based solution typically assumes that the platform can modify or fine-tune the underlying model, whereas many applications access LLMs through APIs and already rely on carefully designed pipelines to ensure response quality. 
In contrast, \pila aims to provide a plug-and-play ad insertion framework that can be attached to existing LLM agentic systems without modifying the base model or disrupting the original harness.

\paragraph{Auction mechanism for LLMs.}
Unlike traditional online advertising auctions that allocate predefined positions or impressions, the value of LLM-native advertising may depend on wording, ordering, and contextual integration. Accordingly, this line of work studies how to design mechanisms that can properly price LLM-generated content.
\citet{duetting2024mechanism} propose a token-level auction framework, where bids can influence the sequential generation process of an LLM. \citet{dubey_auctions_2024} study auctions for LLM-generated summaries, in which advertisers compete for their content to be included in the generated summary. \citet{mordo_sponsored_2024, balseiro2026position} extend position auctions to AI-generated content, where ad effectiveness depends on both the creative and its contextual position. Other studies consider budget constraints \citep{chen_budget-constrained_2024} or preference aggregation during LLM fine-tuning and inference \citep{zhang2025roadmap, sun2026mechanism}. 
However, these works mainly focus on the theoretical analysis and cannot be practically applied to LLM-native ad generation.

\begin{figure*}[!th]
  \centering
  \includegraphics[width=\textwidth]{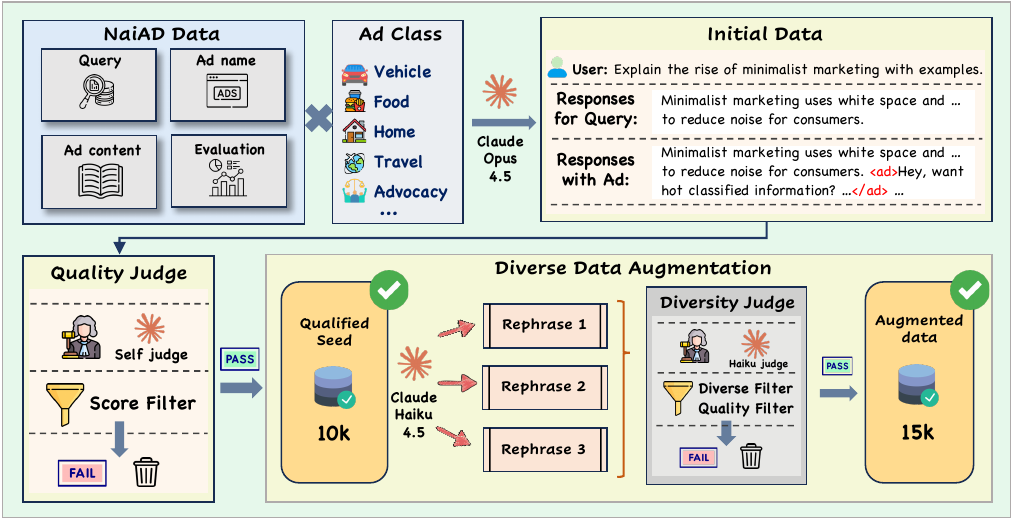}
  \caption{\textbf{The data construction process of the \pila framework.} Starting from NaiAD data covering various ad classes, Claude Opus 4.5 generates initial paired responses (with and without ads) for each query–ad tuple. A quality judge then performs self-judging and score-based filtering to retain a 10k qualified seed set. Each seed is further expanded through diverse data augmentation, where Claude Haiku 4.5 produces three rephrased variants that are passed through a diversity judge applying both diversity and quality filters, ultimately yielding 15k augmented training samples. We combine seed data and augmented data, resulting in 25k high-quality data samples in total.}
  \label{fig:pipeline}
\end{figure*}

\section{Methodology}
\label{sec:method}

In this section, we formally introduce the \pila framework, followed by the data construction process and ad intensity control.

\subsection{\pila: Decoupled Ad Insertion}
\label{sec:task} 

As illustrated in Figure \ref{fig:main}, we cast LLM-native ad insertion as a \emph{conditional response rewriting} problem rather than end-to-end response generation. A request is specified by a tuple $(x,a,c)$, where $x \in \mathcal{X}$ is the user query, and $a$ and $c$ denote the advertiser name and ad content, respectively. A frozen upstream system $\mathcal{M}_{\mathrm{up}}:\mathcal{X}\rightarrow\mathcal{Y}$ first produces an ad-free response
\begin{equation}
y = \mathcal{M}_{\mathrm{up}}(x).
\end{equation}
Here $\mathcal{M}_{\mathrm{up}}$ can be a single LLM or an arbitrary workflow, such as Retrieval-Augmented Generation (RAG) \citep{lewis2020retrieval, arslan2024survey}, tool use \citep{yao2022react, schick2023toolformer}, or multi-agent harness systems \citep{talebirad2023multi}. Importantly, \pila only accesses the final response $y$, without accessing model parameters, intermediate states, or altering the original pipeline.

We instantiate \pila as a sidecar rewriter $\pi_{\theta}$ that takes $(x,y,a,c)$ as input and outputs an ad-enhanced response
\begin{equation}
\tilde{y} \sim \pi_{\theta}(\cdot \mid x, y, a, c),
\label{eq:pila}
\end{equation}
where the inserted sponsored span is enclosed by \texttt{<ad>} and \texttt{</ad>} tags for explicit tracking. This formulation separates answer generation from ad insertion: the upstream system is responsible for solving the user's task, while \pila performs a targeted rewrite on top of the completed answer. Such decoupling is what makes \pila plug-and-play and directly applicable to API-only or workflow-based LLM systems.

\subsection{Data Construction}
\label{sec:data-construction}

Our data construction process is illustrated in Figure \ref{fig:pipeline}. 
We start from user queries collected from public instruction-style corpora and advertisements drawn from a curated brand catalog containing 1,936 ads. Using the NaiAD infrastructure, we convert these 
sources into a unified format and construct supervised training instances of the form $(x, y, a, c, \tilde{y})$. 

We first build a synthesis seed dataset with Claude Opus 4.5 \citep{claude_opus_45}. The synthesis process is carried out in multiple rounds until the target size is reached. In each round, we sample over user queries, and once a query yields an accepted example, it is excluded from subsequent rounds so that later rounds can cover other intents. For each sampled query, we retrieve the top four semantically matched ads using sentence embeddings with a diversity penalty $\lambda{=}0.15$. We then pair the query with each retrieved ad and ask the model to generate both an ad-free response $y$ and an ad-enhanced response $\tilde{y}$, where the sponsored span is explicitly marked with \texttt{<ad>...</ad>}. To encourage variation in insertion style, each candidate is generated under one of four predefined integration strategies. 
Following the NaiAD protocol, the model scores the responses and we retain only high-quality candidates. This stage yields 10,014 accepted seed examples out of 24,883 attempts, while each example is required to lose no more than 2 points from the total score to pass.

To obtain a larger and mode diverse dataset, we further expand the seed set through diversity augmentation with Claude Haiku 4.5 \citep{claude_haiku45}. For each seed example, we generate three paraphrases over the local discourse window containing the \texttt{<ad>} span while preserving the boundary tags and ad content. We keep only variants that introduce meaningful surface variation without reducing quality relative to the original seed. This step yields about 15k additional examples, resulting in a final training set of about 25k instances. We then format the data into the conditional response rewriting structure and fine-tune our \pila models.

\subsection{Controlling Ad Intensity}
\label{sec:ad_intensity_ctrl}

The trained \pila model provides a strong default policy for model-agnostic ad insertion. In practice, different applications may prefer different balances between user-side quality and ad-side exposure: some require advertisements to remain subtle and highly native, while others may value greater salience for visibility, conversion, or downstream monetization. This motivates an inference-time control mechanism that can adjust ad intensity without retraining the \pila model.

Advertising research offers a natural lens for this trade-off. The persuasion knowledge model (PKM) \citep{friestad1994persuasion} suggests that users actively recognize and interpret persuasive intent; when such intent becomes too salient, advertising may undermine trust or perceived relevance. In our setting, stronger insertion can increase the visibility of sponsored content, but may also reduce the naturalness, coherence, or task relevance of the response. Ad intensity should therefore serve as a deployment-time control variable for deciding where a system lies on the spectrum between user experience and advertising utility.

To implement this control, let $\pi_{\pila}$ denote the trained \pila model and $\pi_{\mathrm{ref}}$ the original untuned backbone. Inspired by contrastive decoding \citep{li2023contrastive, mitchell2023emulator, zhang2025amulet}, we define an intensity-controlled policy:
\begin{equation}
\label{eq:cd_pila}
\log \pi_{\pila}^{\rho}
=
\log \pi_{\pila}
+
\rho \log\left(\frac{\pi_{\pila}}{\pi_{\mathrm{ref}}}\right),
\end{equation}
where $\rho$ is the ad-intensity factor. When $\rho=0$, decoding reduces to the standard \pila policy. As $\rho$ increases, the decoder amplifies behaviors learned during \pila training relative to the untuned reference model, producing more salient sponsored content. This gives \pila a simple deployment-time knob for navigating the trade-off between user experience and advertising exposure.

{
\Huge
\setlength{\extrarowheight}{1.4pt}
\setlength{\tabcolsep}{3pt}
\begin{table*}[!t]
\centering
\caption{Main results comparing PILA with single-LLM baselines across four commercial domains. 
We report user-side quality scores (Q1 and Q2), ad-side quality scores (Q3), and their average (Avg). 
Scores range from 1 to 5, where higher is better; the best score in each column is highlighted in bold.}
\label{tab:main_results1}
\resizebox{\textwidth}{!}{%
\begin{tabular}{llccccccccccccccccccc}
\toprule
\multirow{2}{*}{Method} &
  \multirow{2}{*}{Base Model} &
  \multicolumn{4}{c}{Automotive} &
  \multicolumn{1}{c}{} &
  \multicolumn{4}{c}{Food \& Beverage} &
  \multicolumn{1}{c}{} &
  \multicolumn{4}{c}{Home \& Personal Care} &
  \multicolumn{1}{c}{} &
  \multicolumn{4}{c}{Travel} \\
  \cline{3-6} \cline{8-11} \cline{13-16} \cline{18-21}
  &
  &
  \multicolumn{1}{c}{Q1} &
  \multicolumn{1}{c}{Q2} &
  \multicolumn{1}{c}{Q3} &
  \multicolumn{1}{c}{Avg $\uparrow$} &
  \multicolumn{1}{c}{} &
  \multicolumn{1}{c}{Q1} &
  \multicolumn{1}{c}{Q2} &
  \multicolumn{1}{c}{Q3} &
  \multicolumn{1}{c}{Avg $\uparrow$} &
  \multicolumn{1}{c}{} &
  \multicolumn{1}{c}{Q1} &
  \multicolumn{1}{c}{Q2} &
  \multicolumn{1}{c}{Q3} &
  \multicolumn{1}{c}{Avg $\uparrow$} &
  \multicolumn{1}{c}{} &
  \multicolumn{1}{c}{Q1} &
  \multicolumn{1}{c}{Q2} &
  \multicolumn{1}{c}{Q3} &
  \multicolumn{1}{c}{Avg $\uparrow$} \\ \midrule

\multirow{5}{*}[1.3em]{Base}        & Qwen3-4B   &2.02&1.96&3.44&2.47&  &2.00&1.94&3.45&2.46&  &2.04&2.07&3.60&2.57&  &2.12&2.28&3.48&2.63  \\
                                           & Qwen3-8B    &2.00&1.86&3.28&2.38&  &2.03&2.04&3.13&2.40&  &2.07&2.15&3.33&2.52&  &2.11&2.40&3.10&2.54  \\ \cdashline{2-21}
                                           & Average     &2.01&1.91&3.36&2.43&  &2.02&1.99&3.29&2.43&  &2.05&2.11&3.47&2.54&  &2.12&2.34&3.29&2.58  \\ \hline

\multirow{5}{*}[1.3em]{MOSAIC}      & Qwen3-4B   &1.88&1.89&3.38&2.38&  &1.87&1.88&3.23&2.33&  &1.89&1.96&3.44&2.43&  &1.99&2.06&3.35&2.47  \\
                                           & Qwen3-8B    &1.74&1.77&2.96&2.16&  &1.68&1.79&2.81&2.09&  &1.75&1.74&2.88&2.12&  &1.79&1.92&2.88&2.20  \\ \cdashline{2-21}
                                           & Average     &1.81&1.83&3.17&2.27&  &1.78&1.83&3.02&2.21&  &1.82&1.85&3.16&2.28&  &1.89&1.99&3.12&2.33  \\ \hline

\multirow{5}{*}[1.3em]{SFT}         & Qwen3-4B   &2.78&2.67&3.56&3.00&  &2.95&2.75&3.39&3.03&  &3.01&2.76&3.48&3.08&  &2.90&2.95&3.41&3.09  \\
                                           & Qwen3-8B    &2.93&2.90&3.61&3.15&  &2.96&2.81&3.50&3.09&  &2.98&2.89&3.77&3.21&  &3.00&2.99&3.67&3.22  \\ \cdashline{2-21}
                                           & Average     &2.85&2.78&3.59&3.07&  &2.96&2.78&3.45&3.06&  &2.99&2.82&3.63&3.15&  &2.95&2.97&3.54&3.15  \\ \hline

\multirow{5}{*}[1.3em]{\shortstack{PILA-4B \\ \textbf{(ours)}}}     & Qwen3-4B   &3.37&2.82&3.93&3.37&  &3.08&2.78&3.56&3.14&  &3.33&2.92&3.75&3.33&  &3.31&3.14&3.92&3.46  \\
                                           & Qwen3-8B    &3.50&2.91&3.79&3.40&  &3.20&2.84&3.67&3.24&  &3.36&2.88&3.84&3.36&  &3.39&3.08&3.98&3.48  \\ \cdashline{2-21}
                                           & Average     &\textbf{3.43}&2.87&\textbf{3.86}&\textbf{3.39}&  &3.14&2.81&3.61&3.19&  &\textbf{3.34}&2.90&3.80&3.35&  &3.35&3.11&3.95&3.47  \\ \hline

\multirow{5}{*}[1.3em]{\shortstack{PILA-8B \\ \textbf{(ours)}}}     & Qwen3-4B   &3.17&2.89&3.53&3.20&  &3.09&2.87&3.61&3.19&  &3.22&2.90&3.87&3.33&  &3.46&3.16&3.90&3.51  \\
                                           & Qwen3-8B    &3.35&2.96&3.74&3.35&  &3.27&2.86&3.75&3.29&  &3.38&3.01&3.77&3.39&  &3.37&3.24&4.02&3.54  \\ \cdashline{2-21}
                                           & Average     &3.26&\textbf{2.93}&3.63&3.27&  &\textbf{3.18}&\textbf{2.86}&\textbf{3.68}&\textbf{3.24}&  &3.30&\textbf{2.96}&\textbf{3.82}&\textbf{3.36}&  &\textbf{3.41}&\textbf{3.20}&\textbf{3.96}&\textbf{3.52}  \\ \bottomrule

\end{tabular}
}
\end{table*}
}

\section{Experiments}
\label{sec:experiments}

In this section, we conduct extensive experiments to evaluate \pila in various scenarios. 
The results demonstrate that \pila significantly improves the quality of LLM-native advertising in a non-invasive and practical way.

\subsection{Experimental Setups}

We first introduce the experimental setups, including the base and upstream models, the baselines, and the benchmark.

\paragraph{Base Models and Upstream Models.} 
We instantiate \pila using two Qwen3 backbones \citep{yang2025qwen3} of different sizes, 4B and 8B, 
and denote the resulting variants  \pila-4B and \pila-8B, respectively. 
We then evaluate their plug-and-play ad insertion ability, with seven frontier commercial models from major providers as upstream models: GPT-5.4 \citep{gpt54}, GPT-5.4-mini \citep{gpt54_mini}, Gemini 3.1 Pro \citep{gemini31_pro}, Gemini 3.1 Flash \citep{gemini31_flash}, Claude Haiku 4.5 \citep{claude_haiku45}, Deepseek V4 Flash \citep{deepseekai2026deepseekv4}, and Deepseek V3.2 \citep{liu2025deepseek}.

\paragraph{Baseline Methods.}
We compare \pila with several baselines to show its effectiveness over single-LLM approaches:
\begin{itemize}[leftmargin=*]
\item \textbf{Base} refers to prompt-based methods \citep{hajiaghayi_ad_2024, liu2026incentive}. In this paper, we use the ad-insertion prompt from \citet{liu2026incentive}.
\item \textbf{MOSAIC} \citep{soumalias_truthful_2025} samples one response from multiple LLM-generated candidates through an allocation mechanism. For fairness and cost efficiency, we use GPT-5-nano \citep{gpt_5_nano} as the reward backbone and follow the original paper's setting of sampling 20 candidates per query.
\item \textbf{SFT}. Since the only existing tuning-based method, LLM-Auction \citep{zhao2025llm}, uses private data and is not open-sourced, we construct a representative tuning-based baseline by training a single LLM on the high-quality responses from the same data used for \pila.
\end{itemize}

\paragraph{Benchmark and Evaluation.}
NaiAD \citep{zhang2026naiad} is currently the most comprehensive LLM-native advertising dataset, offering broad category coverage and standardized evaluation protocols. 
We build our evaluation benchmark from its six largest advertising categories: Automotive, Food \& Beverage, Home \& Personal Care, Travel, Advocacy, and Financial.
We sample 100 data points from each category with diverse advertiser names and original scores.
Following NaiAD, we use Claude Opus 4.6 \citep{claude_opus46} as the evaluator backbone and retain the prompts and scoring settings.
The evaluator reports three scores that all range from 1 to 5: response relevance (Q1), expression coherence (Q2), and ad effectiveness (Q3). Q1 and Q2 measure user-side quality, and Q3 captures ad-side utility, enabling us to examine the balance of user satisfaction and advertising effectiveness.

\subsection{Experimental Results}
\label{sec:main_exp_results}

\begin{figure}[h]
    \centering
    \includegraphics[width=0.99\linewidth]{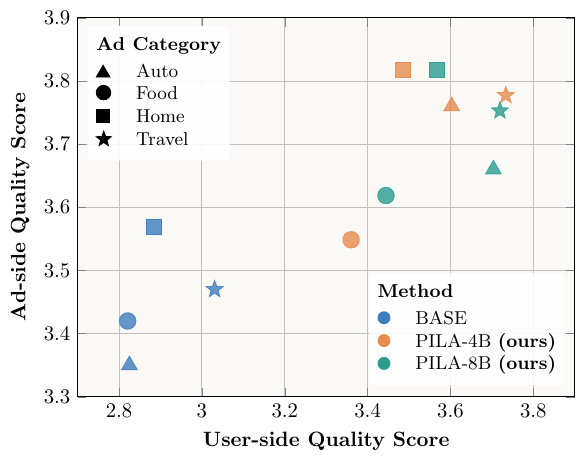}
    \caption{Performance of \pila against base commercial models in terms of user- and ad-side quality across different settings. Each point represents the average result over all evaluated upstream models for a given method and ad category.}
    \label{fig:trade_off_analysis}
\end{figure}
\begin{figure}[h]
    \centering
    \includegraphics[width=0.99\linewidth]{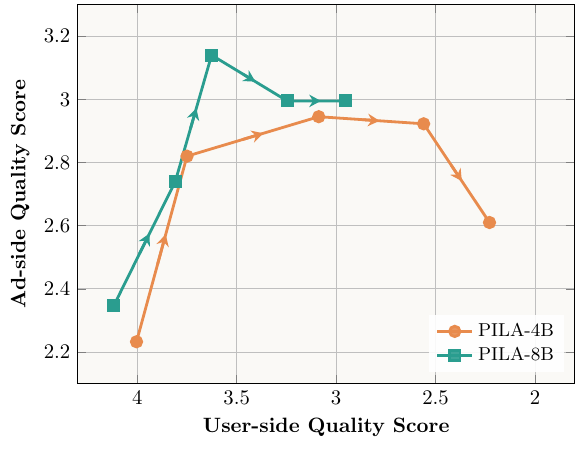}
    \caption{Results of ad intensity control. The x- and y-axes follow the definitions in Figure \ref{fig:trade_off_analysis}, with the x-axis shown in descending order. The arrow indicates the direction of increasing ad intensity factor $\rho$.}
    \label{fig:ad_intensity_ctrl}
\end{figure}

{
\Huge
\setlength{\extrarowheight}{1.4pt}
\setlength{\tabcolsep}{3pt}
\begin{table*}[!t]
\centering
\caption{Comprehensive evaluation of PILA as a plug-and-play enhancement for diverse upstream commercial models. The metrics are the same as those in Table \ref{tab:main_results1}.}
\label{tab:main_results2}
\resizebox{\textwidth}{!}{%
\begin{tabular}{llccccccccccccccccccc}
\toprule
\multirow{2}{*}{Method} &
  \multirow{2}{*}{Upstream Model} &
  \multicolumn{4}{c}{Automotive} &
  \multicolumn{1}{c}{} &
  \multicolumn{4}{c}{Food \& Beverage} &
  \multicolumn{1}{c}{} &
  \multicolumn{4}{c}{Home \& Personal Care} &
  \multicolumn{1}{c}{} &
  \multicolumn{4}{c}{Travel} \\ \cline{3-6} \cline{8-11} \cline{13-16} \cline{18-21} 
  &
  &
  \multicolumn{1}{c}{Q1} &
  \multicolumn{1}{c}{Q2} &
  \multicolumn{1}{c}{Q3} &
  \multicolumn{1}{c}{Avg $\uparrow$} &
  \multicolumn{1}{c}{} &
  \multicolumn{1}{c}{Q1} &
  \multicolumn{1}{c}{Q2} &
  \multicolumn{1}{c}{Q3} &
  \multicolumn{1}{c}{Avg $\uparrow$} &
  \multicolumn{1}{c}{} &
  \multicolumn{1}{c}{Q1} &
  \multicolumn{1}{c}{Q2} &
  \multicolumn{1}{c}{Q3} &
  \multicolumn{1}{c}{Avg $\uparrow$} &
  \multicolumn{1}{c}{} &
 \multicolumn{1}{c}{Q1} &
  \multicolumn{1}{c}{Q2} &
  \multicolumn{1}{c}{Q3} &
  \multicolumn{1}{c}{Avg $\uparrow$} \\ \midrule
\multirow{5}{*}[-2.7em]{Base}        & GPT 5.4   &3.11&2.35&3.19&2.88&  &3.26&2.39&3.33&2.99&  &3.38&2.72&3.32&3.14&  &3.01&2.30&2.87&2.73  \\
                                           & GPT 5.4 mini  &3.41&2.27&3.20&2.96&  &3.08&2.38&3.15&2.87&  &3.29&2.33&3.28&2.97&  &3.41&2.75&3.30&3.15  \\ 
                                           & Gemini 3.1 Pro &3.87&2.85&3.62&3.45&  &3.69&2.84&3.78&3.44&  &3.72&2.85&3.93&3.50&  &3.88&3.23&3.91&3.67  \\
                                           & Gemini 3.1 Flash &2.67&2.38&3.52&2.86&  &2.60&2.23&3.47&2.77&  &2.50&2.19&3.62&2.77&  &2.59&2.58&3.67&2.95  \\
                                           & Claude Haiku 4.5    &2.80&2.68&2.80&2.76&  &2.97&2.78&3.11&2.95&  &3.07&2.96&3.28&3.10&  &3.20&3.21&3.18&3.20  \\
                                           & Deepseek V4 Flash &3.30&2.76&3.57&3.21&  &3.25&2.82&3.56&3.21&  &3.31&2.93&3.91&3.38&  &3.53&3.18&3.64&3.45  \\
                                           & Deepseek V3.2 &2.70&2.40&3.55&2.88&  &2.74&2.46&3.54&2.91&  &2.68&2.45&3.64&2.92&  &2.79&2.77&3.72&3.09  \\ \cdashline{2-21}
                                           & Average     &3.12&2.53&3.35&3.00&  &3.08&2.56&3.42&3.02&  &3.14&2.63&3.57&3.11&  &3.20&2.86&3.47&3.18  \\ \hline
\multirow{5}{*}[-2.7em]{\shortstack{PILA-4B \\ \textbf{(ours)}}}         & GPT 5.4  &3.80&2.96&3.65&3.47&  &3.11&2.50&3.61&3.07&  &3.15&2.61&3.74&3.17&  &3.35&2.75&3.91&3.34  \\
                                           & GPT 5.4 mini   &3.90&2.97&3.63&3.50&  &3.55&2.88&3.35&3.26&  &3.78&2.96&3.83&3.52&  &4.01&3.43&3.65&3.70  \\ 
                                           & Gemini 3.1 Pro  &4.43&3.55&3.73&3.90&  &4.22&3.31&3.52&3.68&  &4.32&3.38&3.78&3.83&  &4.57&3.91&3.77&4.08  \\
                                           & Gemini 3.1 Flash  &4.07&3.30&3.94&3.77&  &4.00&3.18&3.66&3.61&  &4.10&3.34&3.83&3.76&  &4.11&3.41&3.99&3.84  \\
                                           & Claude Haiku 4.5     &3.67&2.85&3.77&3.43&  &3.57&2.77&3.47&3.27&  &3.52&2.87&3.85&3.41&  &3.83&3.29&3.64&3.59  \\
                                           & Deepseek V4 Flash  &4.28&3.52&3.75&3.85&  &4.00&3.21&3.54&3.58&  &4.27&3.46&3.81&3.85&  &4.38&3.78&3.63&3.93  \\
                                           & Deepseek V3.2 &3.96&3.18&3.85&3.66&  &3.69&3.05&3.69&3.48&  &3.92&3.11&3.88&3.64&  &4.06&3.39&3.85&3.77  \\ \cdashline{2-21}
                                           & Average     &4.02&3.19&\textbf{3.76}&3.66&  &3.73&2.99&3.55&3.42&  &3.87&3.10&\textbf{3.82}&3.60&  &\textbf{4.04}&3.42&\textbf{3.78}&\textbf{3.75}  \\ \hline
\multirow{5}{*}[-2.7em]{\shortstack{PILA-8B \\ \textbf{(ours)}}}        & GPT 5.4 &3.84&3.12&3.68&3.55&  &3.46&2.83&3.48&3.26&  &3.61&2.82&3.75&3.39&  &3.92&3.34&3.58&3.61  \\
                                           & GPT 5.4 mini  &4.01&3.17&3.65&3.61&  &3.30&2.87&3.54&3.24&  &3.62&2.77&3.67&3.35&  &3.62&3.16&3.93&3.57  \\ 
                                           & Gemini 3.1 Pro &4.50&3.70&3.74&3.98&  &3.98&3.17&3.62&3.59&  &4.33&3.40&3.77&3.83&  &4.19&3.58&3.91&3.89  \\
                                           & Gemini 3.1 Flash &4.28&3.66&3.74&3.89&  &3.99&3.28&3.99&3.75&  &3.97&3.45&4.00&3.81&  &3.90&3.50&3.60&3.67  \\
                                           & Claude Haiku 4.5    &3.64&2.98&3.60&3.41&  &3.75&2.95&3.50&3.40&  &3.70&3.04&3.85&3.53&  &3.59&3.21&3.82&3.54  \\
                                           & Deepseek V4 Flash &4.13&3.39&3.60&3.71&  &4.01&3.33&3.58&3.64&  &4.18&3.48&3.85&3.84&  &4.39&3.87&3.62&3.96  \\
                                           & Deepseek V3.2 &4.09&3.34&3.61&3.68&  &4.01&3.29&3.62&3.64&  &4.17&3.40&3.83&3.80&  &4.17&3.63&3.81&3.87  \\ \cdashline{2-21}
                                           & Average     &\textbf{4.07}&\textbf{3.34}&3.66&\textbf{3.69}&  &\textbf{3.79}&\textbf{3.10}&\textbf{3.62}&\textbf{3.50}&  &\textbf{3.94}&\textbf{3.19}&\textbf{3.82}&\textbf{3.65}&  &3.97&\textbf{3.47}&3.75&3.73  \\ \bottomrule
\end{tabular}
}
\end{table*}
}

To thoroughly and systematically evaluate \pila's performance, we conduct our main experiments under three settings.
First, we compare \pila against single-LLM baselines to show that \pila offers clear advantages over methods that integrate content generation and ad insertion within one model.
Second, we examine whether \pila can enhance various commercial models in a plug-and-play manner, and show that it substantially improves ad-insertion quality without degrading the original models' ability to generate high-quality user responses.
We also demonstrate the great potential of our ad-intensity control method.
Due to space limitations, we present the results on the first four evaluation task categories in the main text, while the results on the remaining two categories are reported in the appendix.
We introduce the results of these experiments in the following paragraphs.

\paragraph{Results on Single LLM Settings.}
We compare our method with single-LLM baselines. 
As shown in Table \ref{tab:main_results1}, our method achieves the best overall performance across all settings. 
In terms of the average score, \pila-8B performs best in three out of four scenarios, while \pila-4B accounts for
the remaining one. SFT ranks second, followed by Base and MOSAIC. Across the four ad-insertion tasks, \pila-8B
improves over SFT, Base, and MOSAIC by 7.7\%, 34.2\%, and 47.3\% on average, respectively, while \pila-4B achieves
corresponding gains of 7.8\%, 34.3\%, and 47.4\%.
Although \pila-8B performs better across a broader range of settings, \pila-4B obtains a notably higher score on Automotive, making its average improvement close to that of the 8B model.
These results demonstrate the effectiveness of our method over single-model methods in improving both user-side and ad-side quality for LLM-native advertising.

\paragraph{Results on Improving Upstream Models.}
We then analyze our core experiments: how much improvement \pila can bring to upstream commercial models. Table \ref{tab:main_results2} shows that both \pila-4B and \pila-8B consistently improve all scores across all frontier upstream models and ad topics, achieving the best performance in all scenarios and demonstrating the effectiveness of our method.

On average, \pila-4B improves Q1, Q2, Q3, and Avg by 24.9\%, 20.0\%, 8.1\%, and 17.2\%, respectively.
\pila-8B yields even stronger overall gains, with 25.8\%, 23.8\%, 7.5\%, and 18.4\%, respectively.
The gains are especially pronounced for several strong upstream models: \pila-8B improves Avg by 33.2\% on Gemini
3.1 Flash, 27.0\% on Deepseek V3.2, and 17.6\% on GPT 5.4.
These results show that \pila can serve as a general model-agnostic module for enhancing a wide range of
upstream models.

We further analyze the ability to balance user-side and ad-side objectives.
The results are shown in Figure \ref{fig:trade_off_analysis}. The x-axis denotes the user-side score, computed as the average of Q1 and Q2, while the y-axis denotes the ad-side score Q3. 
We observe that our method achieves an almost Pareto improvement over the Base method. Across all experimental settings, \pila consistently appears in the upper-right region relative to Base. This suggests that, compared with direct ad-aware generation, the two-stage generation paradigm better preserves the upstream model’s ability to satisfy user needs, while steadily shifting it toward a more favorable trade-off region where ad-side quality improves without degrading user-side satisfaction.

\paragraph{Results on Ad Intensity Control.}
As described in methodology, to satisfy pluralistic real-world trade-off requirements between the user and the ad side, 
we design an ad intensity control method that combines PKM and contrastive decoding.
For this evaluation, we randomly sample 10\% of queries and their corresponding ad data for each ad topic. For each example, we further evaluate five different settings with $\rho \in \{0,1,2,3,4\}$. The results are shown in Figure \ref{fig:ad_intensity_ctrl}.
The definitions of the x- and y-axes are the same as in Figure \ref{fig:trade_off_analysis}. 
The only difference is that the x-axis is shown in descending order. Each point represents the expected user-side and ad-side scores under a specific value of the ad-intensity factor $\rho$, and the arrows along the curve indicate the direction in which $\rho$ increases.

From the figure, we observe that for both \pila-4B and \pila-8B, the user-side score decreases monotonically as $\rho$ increases. This reflects the PKM-based intuition that overly strengthening advertising can reduce users’ perceived satisfaction. In contrast, the ad-side score first increases and then decreases, suggesting that moderate ad enhancement can improve ad insertion quality, while excessive emphasis on the ad may also degrade it to some extent. Nevertheless, the ad-side scores remain higher than the $\rho = 0$ setting, indicating that they still outperform the case without any ad intensity enhancement.

\subsection{Case Study}
\label{sec:case_study}

To demonstrate how \pila improves ad-insertion quality in practice, we present a qualitative case study comparing it against the baseline. This comparison also serves to validate that our automated evaluation metrics (Q1 to Q3) accurately reflect human-perceived quality.
Table \ref{tab:case_qualitative} illustrates a paired example using the same user query (about liver qi stagnation) and ad info (\textsc{Liverite}).
The \textbf{Base} method 
receives lower scores across the board (Q1/Q2/Q3 of 3.5/3.2/3.2). While it successfully provides the pathomechanism, the response abruptly pivots from a Traditional Chinese Medicine exposition directly into detox marketing. This sudden shift compromises both relevance and coherence (Q1, Q2), making the ad feel artificially pasted rather than naturally integrated (Q3).
In contrast, \textbf{PILA} rewrites the completed answer to achieve significantly higher scores (4.2/4.2/4.3). It leaves the clinical explanation intact but introduces a dedicated bridging section before the \texttt{<ad>} tag. By tying the ad content to a shared narrative about liver function, \pila seamlessly weaves the ad into the context.
Ultimately, the gap in these automated scores clearly aligns with the human reading experience, reinforcing the aggregate gains reported in Table \ref{tab:main_results2}. Additional cases are available in the case study section of the appendix.

\begin{table}[t]
\centering
\footnotesize
\setlength{\tabcolsep}{5pt}
\renewcommand{\arraystretch}{1.12}
\caption{Qualitative comparison of ad-insertion methods on a fixed query--ad pair. 
Notice the abrupt, unnatural pivot in the {\color{violet}\textbf{Base method }} compared to {\color{orange}\textbf{\pila's smooth semantic hooks}}. }
\label{tab:case_qualitative}
\begin{tabularx}{\linewidth}{@{}>{\centering\arraybackslash}m{0.9cm} >{\raggedright\arraybackslash}X@{}}
\toprule
\multicolumn{2}{@{}p{\linewidth}@{}}{\textbf{Query:} Explain the pathomechanism of liver qi stagnation} \\[4pt]
\multicolumn{2}{@{}p{\linewidth}@{}}{\textbf{Ad (\textsc{Liverite}):} \ldots{} Liverite is the all-natural detoxifier and cleanser that helps your liver cope with toxins, medications, and alcohol. Your liver is the key to better health and energy.} \\
\midrule
\textbf{Method} & \textbf{Insertion excerpt} \\
\midrule
Base & In traditional Chinese medicine, liver qi stagnation refers to a pattern where the Liver's normal function of promoting the smooth flow of qi becomes impaired. The Liver is said to ensure the free movement of qi, emotions, blood, and body fluids throughout the body... \textit{[... detailed pathomechanism omitted ...]} \newline\vspace*{-2mm}\newline
In practical terms, many people notice that when they feel sluggish, run down, or ``not themselves,'' they associate it with poor liver function. {\bfseries\ttfamily\textless ad\textgreater}{\itshape {\color{violet}\textbf{That's why products like Liverite are marketed as}} an all-natural detoxifier and cleanser---promoting the idea of helping the liver cope with toxins, medications, alcohol, and everyday environmental stressors, and supporting healthy liver function and energy.}{\bfseries\ttfamily\textless/ad\textgreater} \\
\midrule
PILA & In Traditional Chinese Medicine, liver qi stagnation refers to a pattern where the Liver's normal function of ensuring the smooth flow of qi is disrupted. The pathomechanism is usually described like this... \textit{[... detailed pathomechanism omitted ...]} \newline\vspace*{-2mm}\newline
\textbf{Bridging TCM and Western physiology} \newline
It's fascinating how {\color{orange}\textbf{both systems converge}} on the same insight: the liver isn't just a passive filter---it's the body's command center for energy, detoxification, and emotional regulation. {\bfseries\ttfamily\textless ad\textgreater}{\itshape {\color{orange}\textbf{Western medicine recognizes this too}}: your liver processes everything from last night's drink to environmental pollutants, and when it's overwhelmed, you feel it---run-down, sluggish, like something's grinding you down internally. Liverite works with this reality, delivering an all-natural formula that supports your liver's role as the body's main filter. 
}{\bfseries\ttfamily\textless/ad\textgreater} \\
\bottomrule
\end{tabularx}
\end{table}

\section{Conclusion}
\label{sec:conclusion}

In this paper, we presented \pila, a plug-and-play sidecar framework for LLM-native advertising that decouples ad insertion from upstream response generation. By formulating ad insertion as a conditional response rewriting problem, \pila can be attached to arbitrary LLM services without modifying the base model, making it compatible with API-only systems, agentic pipelines, and existing production stacks.

Our experiments in various scenarios show that \pila consistently improves ad-side effectiveness while preserving user-side response satisfaction. Beyond average performance gains, the proposed intensity controller further exposes a smooth and controllable trade-off between naturalness and exposure, which makes the framework more practical for real-world deployment and downstream pricing mechanisms. 
These results highlight decoupled ad insertion as a practical path for general-purpose, model-agnostic, controllable, and deployable LLM-native advertising in real-world applications.

\bibliography{reference}

@article{zhao2025llm,
  title={LLM-Auction: Generative Auction towards LLM-Native Advertising},
  author={Zhao, Chujie and Hu, Qun and Song, Shiping and Chen, Dagui and Zhu, Han and Xu, Jian and Zheng, Bo},
  journal={arXiv preprint arXiv:2512.10551},
  year={2025}
}

@article{wu2025generative,
  title={What generative search engines like and how to optimize web content cooperatively},
  author={Wu, Yujiang and Zhong, Shanshan and Kim, Yubin and Xiong, Chenyan},
  journal={arXiv preprint arXiv:2510.11438},
  year={2025}
}

@article{liu2025deepseek,
  title={Deepseek-v3.2: Pushing the frontier of open large language models},
  author={Liu, Aixin and Mei, Aoxue and Lin, Bangcai and Xue, Bing and Wang, Bingxuan and Xu, Bingzheng and Wu, Bochao and Zhang, Bowei and Lin, Chaofan and Dong, Chen and others},
  journal={arXiv preprint arXiv:2512.02556},
  year={2025}
}

@article{talebirad2023multi,
  title={Multi-agent collaboration: Harnessing the power of intelligent llm agents},
  author={Talebirad, Yashar and Nadiri, Amirhossein},
  journal={arXiv preprint arXiv:2306.03314},
  year={2023}
}

@article{schick2023toolformer,
  title={Toolformer: Language models can teach themselves to use tools},
  author={Schick, Timo and Dwivedi-Yu, Jane and Dess{\`\i}, Roberto and Raileanu, Roberta and Lomeli, Maria and Hambro, Eric and Zettlemoyer, Luke and Cancedda, Nicola and Scialom, Thomas},
  journal={Advances in neural information processing systems},
  volume={36},
  pages={68539--68551},
  year={2023}
}

@article{sun2026mechanism,
  title={Mechanism design for llm fine-tuning with multiple reward models},
  author={Sun, Haoran and Chen, Yurong and Wang, Siwei and Xu, Chu and Chen, Wei and Deng, Xiaotie},
  journal={Advances in Neural Information Processing Systems},
  volume={38},
  pages={159692--159734},
  year={2026}
}

@article{lewis2020retrieval,
  title={Retrieval-augmented generation for knowledge-intensive nlp tasks},
  author={Lewis, Patrick and Perez, Ethan and Piktus, Aleksandra and Petroni, Fabio and Karpukhin, Vladimir and Goyal, Naman and K{\"u}ttler, Heinrich and Lewis, Mike and Yih, Wen-tau and Rockt{\"a}schel, Tim and others},
  journal={Advances in neural information processing systems},
  volume={33},
  pages={9459--9474},
  year={2020}
}

@article{zhang2026naiad,
  title={NaiAD: Initiate Data-Driven Research for LLM Advertising},
  author={Zhang, Yihang and Huang, Zimeng and Zhai, Ren and Kang, Yipeng and Wang, Tonghan},
  journal={arXiv preprint arXiv:2605.09918},
  year={2026}
}

@inproceedings{duetting2024mechanism,
  title={Mechanism design for large language models},
  author={Duetting, Paul and Mirrokni, Vahab and Paes Leme, Renato and Xu, Haifeng and Zuo, Song},
  booktitle={Proceedings of the ACM Web Conference 2024},
  pages={144--155},
  year={2024}
}

@inproceedings{chen_budget-constrained_2024,
	address = {Singapore Singapore},
	title = {Budget-{Constrained} {Auctions} with {Unassured} {Priors}: {Strategic} {Equivalence} and {Structural} {Properties}},
	isbn = {979-8-4007-0171-9},
	shorttitle = {Budget-{Constrained} {Auctions} with {Unassured} {Priors}},
	url = {https://dl.acm.org/doi/10.1145/3589334.3645344},
	doi = {10.1145/3589334.3645344},
	language = {en},
	urldate = {2026-05-03},
	booktitle = {Proceedings of the {ACM} {Web} {Conference} 2024},
	publisher = {ACM},
	author = {Chen, Zhaohua and Yang, Mingwei and Wang, Chang and Li, Jicheng and Cai, Zheng and Ren, Yukun and Zhu, Zhihua and Deng, Xiaotie},
	month = may,
	year = {2024},
	pages = {14--24}
}

@inproceedings{aggarwal2024geo,
  title={Geo: Generative engine optimization},
  author={Aggarwal, Pranjal and Murahari, Vishvak and Rajpurohit, Tanmay and Kalyan, Ashwin and Narasimhan, Karthik and Deshpande, Ameet},
  booktitle={Proceedings of the 30th ACM SIGKDD conference on knowledge discovery and data mining},
  pages={5--16},
  year={2024}
}

@misc{dubey_auctions_2024,
	title = {Auctions with {LLM} {Summaries}},
	url = {http://arxiv.org/abs/2404.08126},
	doi = {10.48550/arXiv.2404.08126},
	language = {en},
	urldate = {2026-05-03},
	publisher = {arXiv},
	author = {Dubey, Kumar Avinava and Feng, Zhe and Kidambi, Rahul and Mehta, Aranyak and Wang, Di},
	month = apr,
	year = {2024},
	note = {arXiv:2404.08126 [cs]}
}

@inproceedings{hajiaghayi_ad_2024,
	address = {Vancouver, BC, Canada},
	title = {Ad {Auctions} for {LLMs} via {Retrieval} {Augmented} {Generation}},
	isbn = {979-8-3313-1438-5},
	url = {http://www.proceedings.com/079017-0585.html},
	doi = {10.52202/079017-0585},
	language = {en},
	urldate = {2026-05-03},
	booktitle = {Advances in {Neural} {Information} {Processing} {Systems} 37},
	publisher = {Neural Information Processing Systems Foundation, Inc. (NeurIPS)},
	author = {Hajiaghayi, MohammadTaghi and Lahaie, Sébastien and Rezaei, Keivan and Shin, Suho},
	year = {2024},
	pages = {18445--18480},
}

@inproceedings{balseiro2026position,
  title={Position auctions in ai-generated content},
  author={Balseiro, Santiago and Bhawalkar, Kshipra and Deng, Yuan and Feng, Zhe and Mao, Jieming and Mehta, Aranyak and Mirrokni, Vahab and Paes Leme, Renato and Wang, Di and Zuo, Song},
  booktitle={Proceedings of the ACM Web Conference 2026},
  pages={261--272},
  year={2026}
}

@inproceedings{zhang2025roadmap,
  title={Roadmap on incentive compatibility for ai alignment and governance in sociotechnical systems},
  author={Zhang, Zhaowei and Bai, Fengshuo and Wang, Mingzhi and Ye, Haoyang and Ma, Chengdong and Yang, Yaodong},
  booktitle={International Conference on Artificial General Intelligence},
  pages={370--380},
  year={2025},
  organization={Springer}
}

@article{liu2026incentive,
  title={Incentive-Aware Multi-Fidelity Optimization for Generative Advertising in Large Language Models},
  author={Liu, Jiayuan and Wang, Barry and Gan, Jiarui and Wang, Tonghan and Xie, Leon and Guo, Mingyu and Conitzer, Vincent},
  journal={arXiv preprint arXiv:2604.06263},
  year={2026}
}

@article{yun2026llm,
  title={LLM Advertisement based on Neuron Auctions},
  author={Yun, Peiran and Xu, Wenxin and Liu, Jiayuan and Zhang, Yihang and Zeng, Liang and Kong, Lingkai and Wang, Tonghan},
  journal={arXiv preprint arXiv:2605.08326},
  year={2026}
}

@inproceedings{mordo_sponsored_2024,
	title = {Sponsored {Question} {Answering}},
	url = {http://arxiv.org/abs/2407.04471},
	doi = {10.1145/3664190.3672517},
	language = {en},
	urldate = {2026-05-03},
	booktitle = {Proceedings of the 2024 {ACM} {SIGIR} {International} {Conference} on {Theory} of {Information} {Retrieval}},
	author = {Mordo, Tommy and Tennenholtz, Moshe and Kurland, Oren},
	month = aug,
	year = {2024},
	note = {arXiv:2407.04471 [cs]},
	pages = {167--173}
}

@misc{soumalias_truthful_2025,
	title = {Truthful {Aggregation} of {LLMs} with an {Application} to {Online} {Advertising}},
	url = {http://arxiv.org/abs/2405.05905},
	doi = {10.48550/arXiv.2405.05905},
	language = {en},
	urldate = {2026-05-03},
	publisher = {arXiv},
	author = {Soumalias, Ermis and Curry, Michael J. and Seuken, Sven},
	month = feb,
	year = {2025},
	note = {arXiv:2405.05905 [cs]}
}

@article{xu2026ad,
  title={Ad Insertion in LLM-Generated Responses},
  author={Xu, Shengwei and Chen, Zhaohua and Deng, Xiaotie and Huang, Zhiyi and Schoenebeck, Grant},
  journal={arXiv preprint arXiv:2601.19435},
  year={2026}
}

@article{achiam2023gpt,
  title={Gpt-4 technical report},
  author={Achiam, Josh and Adler, Steven and Agarwal, Sandhini and Ahmad, Lama and Akkaya, Ilge and Aleman, Florencia Leoni and Almeida, Diogo and Altenschmidt, Janko and Altman, Sam and Anadkat, Shyamal and others},
  journal={arXiv preprint arXiv:2303.08774},
  year={2023}
}

@misc{observer2026openaiads,
  author       = {Spivack, Jameson}, 
  title        = {The Problem With {OpenAI} Putting Ads in {ChatGPT}},
  howpublished = {Observer},
  year         = {2026},
  month        = {January 23},
  url          = {https://observer.com/2026/01/the-problem-with-openai-putting-ads-in-chatgpt/},
  note         = {Accessed: May 4, 2026}
}

@misc{ndtv2026backlash,
  author       = {{NDTV News Desk}},
  title        = {{OpenAI} Faces Backlash Over Ads Appearing in {ChatGPT}, Users Advise "Don't Do It"},
  howpublished = {NDTV},
  year         = {2025},
  month        = {December 5},
  url          = {https://www.ndtv.com/feature/openai-faces-backlash-over-ads-appearing-in-chatgpt-users-advise-dont-do-it-9754275},
  note         = {Accessed: May 4, 2026}
}

@misc{openai_ad,
  author = {OpenAI},
  title = {Ad policies},
  year = {2026},
  note = {Updated: April 29, 2026},
  url = {https://openai.com/policies/ad-policies/}
}

@misc{claude_opus46,
  author = {Anthropic},
  title = {Introducing Claude Opus 4.6},
  year = {2026},
  note = {Updated: February 5, 2026},
  url = {https://www.anthropic.com/news/claude-opus-4-6}
}

@misc{claude_haiku45,
  author = {Anthropic},
  title = {Introducing Claude Haiku 4.5},
  year = {2025},
  note = {Updated: October 15, 2025},
  url = {https://www.anthropic.com/news/claude-haiku-4-5}
}

@misc{gpt54,
  author = {OpenAI},
  title = {Introducing GPT‑5.4},
  year = {2026},
  note = {Updated: March 5, 2026},
  url = {https://openai.com/index/introducing-gpt-5-4}
}

@misc{gpt54_mini,
  author = {OpenAI},
  title = {Introducing GPT‑5.4 mini and nano},
  year = {2026},
  note = {Updated: March 17, 2026},
  url = {https://openai.com/index/introducing-gpt-5-4-mini-and-nano}
}

@misc{gpt_5_nano,
  author = {OpenAI Developers},
  title = {GPT-5 nano},
  year = {2025},
  url = {https://developers.openai.com/api/docs/models/gpt-5-nano}
}

@misc{claude_opus_45,
  author = {Anthropic},
  title = {Introducing Claude Opus 4.5},
  year = {2025},
  url = {https://www.anthropic.com/news/claude-opus-4-5}
}

@misc{gemini31_pro,
  author = {Google DeepMind},
  title = {Gemini 3.1 Pro},
  year = {2026},
  note = {Updated: February 19, 2026},
  url = {https://deepmind.google/models/model-cards/gemini-3-1-pro}
}

@misc{gemini31_flash,
  author = {Google DeepMind},
  title = {Introducing 3.1 Flash-Lite, a scalable thinking model for high-volume tasks at low cost and latency},
  year = {2026},
  url = {https://deepmind.google/models/gemini/flash-lite}
}

@misc{deepseekai2026deepseekv4,
      title={DeepSeek-V4: Towards Highly Efficient Million-Token Context Intelligence},
      author={DeepSeek-AI},
      year={2026},
}

@inproceedings{zhuge2024gptswarm,
  title={Gptswarm: Language agents as optimizable graphs},
  author={Zhuge, Mingchen and Wang, Wenyi and Kirsch, Louis and Faccio, Francesco and Khizbullin, Dmitrii and Schmidhuber, J{\"u}rgen},
  booktitle={Forty-first International Conference on Machine Learning},
  year={2024}
}

@inproceedings{zhang2025aflow,
  title={Aflow: Automating agentic workflow generation},
  author={Zhang, Jiayi and Xiang, Jinyu and Yu, Zhaoyang and Teng, Fengwei and Chen, Xionghui and Chen, Jiaqi and Zhuge, Mingchen and Cheng, Xin and Hong, Sirui and Wang, Jinlin and others},
  booktitle={International Conference on Learning Representations},
  volume={2025},
  pages={34040--34077},
  year={2025}
}

@inproceedings{zhang2025amulet,
  title={Amulet: Realignment during test time for personalized preference adaptation of LLMs},
  author={Zhang, Zhaowei and Bai, Fengshuo and Chen, Qizhi and Ma, Chengdong and Wang, Mingzhi and Sun, Haoran and Zheng, Zilong and Yang, Yaodong},
  booktitle={International Conference on Learning Representations},
  volume={2025},
  pages={74039--74069},
  year={2025}
}

@inproceedings{li2023contrastive,
  title={Contrastive decoding: Open-ended text generation as optimization},
  author={Li, Xiang Lisa and Holtzman, Ari and Fried, Daniel and Liang, Percy and Eisner, Jason and Hashimoto, Tatsunori B and Zettlemoyer, Luke and Lewis, Mike},
  booktitle={Proceedings of the 61st annual meeting of the association for computational linguistics (volume 1: Long papers)},
  pages={12286--12312},
  year={2023}
}

@article{mitchell2023emulator,
  title={An emulator for fine-tuning large language models using small language models},
  author={Mitchell, Eric and Rafailov, Rafael and Sharma, Archit and Finn, Chelsea and Manning, Christopher D},
  journal={arXiv preprint arXiv:2310.12962},
  year={2023}
}

@inproceedings{burns2016design,
  title={Design patterns for container-based distributed systems},
  author={Burns, Brendan and Oppenheimer, David},
  booktitle={8th USENIX Workshop on Hot Topics in Cloud Computing (HotCloud 16)},
  year={2016}
}

@article{friestad1994persuasion,
  title={The persuasion knowledge model: How people cope with persuasion attempts},
  author={Friestad, Marian and Wright, Peter},
  journal={Journal of consumer research},
  volume={21},
  number={1},
  pages={1--31},
  year={1994},
  publisher={The University of Chicago Press}
}

@article{yang2025qwen3,
  title={Qwen3 technical report},
  author={Yang, An and Li, Anfeng and Yang, Baosong and Zhang, Beichen and Hui, Binyuan and Zheng, Bo and Yu, Bowen and Gao, Chang and Huang, Chengen and Lv, Chenxu and others},
  journal={arXiv preprint arXiv:2505.09388},
  year={2025}
}

@misc{claudecode,
  title = {Claude Code documentation},
  author = {Anthropic},
  year = {2025},
  url = {https://code.claude.com/docs/en/overview}
}

@misc{agent_harness,
  title = {Harness design for long-running application development},
  author = {Anthropic},
  year = {2026},
  url = {https://www.anthropic.com/engineering/harness-design-long-running-apps}
}

@misc{harness_engineering,
  title = {Harness engineering: leveraging Codex in an agent-first world},
  author = {Ryan Lopopolo},
  year = {2026},
  url = {https://openai.com/index/harness-engineering}
}

@article{yao2022react,
  title={React: Synergizing reasoning and acting in language models},
  author={Yao, Shunyu and Zhao, Jeffrey and Yu, Dian and Du, Nan and Shafran, Izhak and Narasimhan, Karthik and Cao, Yuan},
  journal={arXiv preprint arXiv:2210.03629},
  year={2022}
}

@article{arslan2024survey,
  title={A Survey on RAG with LLMs},
  author={Arslan, Muhammad and Ghanem, Hussam and Munawar, Saba and Cruz, Christophe},
  journal={Procedia computer science},
  volume={246},
  pages={3781--3790},
  year={2024},
  publisher={Elsevier}
}

@misc{codex,
  title = {Codex documentation},
  author = {OpenAI},
  year = {2025},
  url = {https://openai.com/codex}
}

@misc{openclaw,
  title = {OpenClaw documentation},
  author = {OpenClaw},
  year = {2026},
  url = {https://docs.openclaw.ai}
}

@inproceedings{zheng2024llamafactory,
  title={Llamafactory: Unified efficient fine-tuning of 100+ language models},
  author={Zheng, Yaowei and Zhang, Richong and Zhang, Junhao and Ye, Yanhan and Luo, Zheyan},
  booktitle={Proceedings of the 62nd annual meeting of the association for computational linguistics (volume 3: system demonstrations)},
  pages={400--410},
  year={2024}
}

@misc{alpaca,
  author = {Rohan Taori and Ishaan Gulrajani and Tianyi Zhang and Yann Dubois and Xuechen Li and Carlos Guestrin and Percy Liang and Tatsunori B. Hashimoto },
  title = {Stanford Alpaca: An Instruction-following LLaMA model},
  year = {2023},
  publisher = {GitHub},
  journal = {GitHub repository},
  howpublished = {\url{https://github.com/tatsu-lab/stanford_alpaca}},
}





\onecolumn

\appendix



\section{LLM-native Advertising}
\label{app:show_cases}

In this section, we present a concrete example of LLM-native advertising. Unlike a separate banner or appended promotional block, the advertisement is generated as part of the model's answer and placed inside a contextually relevant reasoning segment.
In this example, the user asks about protein absorption, and the model naturally connects the advertised product to the discussion of cooked and convenient protein sources. As a result, the advertisement, highlighted in red, follows the structure, tone, and informational flow of the LLM response while remaining embedded in the generated text.

\begin{tcolorbox}[breakable]
\textbf{User Query:} How to improve protein absorption?

\textbf{Advertisement:} Tyson Beef Tips

\textbf{Generated Response:}

Improving protein absorption involves optimizing both dietary choices and digestive
health. In general, it helps to spread protein intake across several meals, choose
digestible protein sources, cook protein properly, support gut health, stay hydrated,
and maintain adequate vitamin and mineral intake.

For example, cooked proteins are often easier to digest than raw or undercooked
sources, since heat can help break down protein structures and connective tissue.\textcolor{red}{
\textless ad\textgreater
For a convenient cooked protein option, Tyson Beef Tips in Gravy are pre-seasoned,
slow-cooked, and ready in five minutes. They provide a quick protein source that can
fit into a balanced meal routine.
\textless /ad\textgreater}

Other useful strategies include pairing protein with moderate carbohydrates, avoiding
excessive alcohol, and addressing any underlying digestive conditions with help from a
healthcare professional. Overall, the best approach is a balanced diet that combines
high-quality protein sources with good digestive health habits.

\textbf{Evaluation Scores:} q1 = 5.0, q2 = 4.2, q3 = 4.5
\end{tcolorbox}



\section{Supplementary Experiments}
\label{app:supp_experiments}

In this section, we will provide more details about our experiments.

{
\Huge
\setlength{\extrarowheight}{1.4pt}
\setlength{\tabcolsep}{3pt}
\begin{table*}[ht]
\centering
\caption{Main results on the remaining two commercial domains for the comparison with single-LLM baselines. We report scores on user-side quality metrics (Q1, Q2), ad-side quality metric (Q3), and their average score (Avg). The best score in each column within each domain is highlighted in bold.}
\label{tab:main_results_complement1}
\resizebox{0.65\textwidth}{!}{%
\begin{tabular}{llccccccccc}
\toprule
\multirow{2}{*}{Method} &
  \multirow{2}{*}{Base Model} &
  \multicolumn{4}{c}{Financial} &
  \multicolumn{1}{c}{} &
  \multicolumn{4}{c}{Beauty} \\
  \cline{3-6} \cline{8-11}
  &
  &
  \multicolumn{1}{c}{Q1} &
  \multicolumn{1}{c}{Q2} &
  \multicolumn{1}{c}{Q3} &
  \multicolumn{1}{c}{Avg $\uparrow$} &
  \multicolumn{1}{c}{} &
  \multicolumn{1}{c}{Q1} &
  \multicolumn{1}{c}{Q2} &
  \multicolumn{1}{c}{Q3} &
  \multicolumn{1}{c}{Avg $\uparrow$} \\ \midrule
\multirow{5}{*}[1.3em]{Base}        & Qwen3-4B   &2.09&2.46&3.35&2.63&  &2.01&2.09&3.44&2.51  \\
                                           & Qwen3-8B   &2.21&2.48&2.99&2.56&  &2.25&2.20&3.14&2.53  \\ \cdashline{2-11}
                                           & Average     &2.15&2.47&3.17&2.60&  &2.13&2.14&3.29&2.52  \\ \hline
\multirow{5}{*}[1.3em]{MOSAIC}        & Qwen3-4B   &1.91&2.29&3.13&2.44&  &1.90&1.97&3.29&2.39  \\
                                           & Qwen3-8B   &1.82&2.07&2.72&2.20&  &2.07&1.90&2.82&2.26  \\ \cdashline{2-11}
                                           & Average     &1.86&2.18&2.93&2.32&  &1.98&1.93&3.06&2.32  \\ \hline
\multirow{5}{*}[1.3em]{SFT}        & Qwen3-4B   &3.04&3.06&3.31&3.14&  &3.26&3.01&3.41&3.23  \\
                                           & Qwen3-8B   &3.23&3.04&3.50&3.26&  &3.25&2.95&3.64&3.28  \\ \cdashline{2-11}
                                           & Average     &3.14&3.05&3.40&3.20&  &3.25&2.98&3.53&3.25  \\ \hline
\multirow{5}{*}[1.3em]{\shortstack{PILA-4B \\ \textbf{(ours)}}}        & Qwen3-4B   &3.25&3.08&3.63&3.32&  &3.58&3.07&3.74&3.46  \\
                                           & Qwen3-8B   &3.27&2.99&3.52&3.26&  &3.62&3.05&3.89&3.52  \\ \cdashline{2-11}
                                           & Average     &3.26&3.04&3.58&3.29&  &\textbf{3.60}&3.06&\textbf{3.82}&\textbf{3.49}  \\ \hline
\multirow{5}{*}[1.3em]{\shortstack{PILA-8B \\ \textbf{(ours)}}}       & Qwen3-4B   &3.32&3.19&3.62&3.38&  &3.58&3.15&3.81&3.51  \\
                                           & Qwen3-8B   &3.29&3.06&3.62&3.32&  &3.52&3.10&3.81&3.48  \\ \cdashline{2-11}
                                           & Average     &\textbf{3.31}&\textbf{3.13}&\textbf{3.62}&\textbf{3.35}&  &3.55&\textbf{3.12}&3.81&\textbf{3.49}  \\ \bottomrule
\end{tabular}
}
\end{table*}
}

\subsection{Supplementary Main Results}
\label{app:full_main_results}
This subsection reports the remaining two commercial domains that are not shown in the main text due to space limitations. We follow the same setup as in Section \ref{sec:experiments} and report Q1, Q2, Q3, and Avg. Here Q1 and Q2 measure user-side response quality, while Q3 measures ad-side effectiveness.

{
\Huge
\setlength{\extrarowheight}{1.4pt}
\setlength{\tabcolsep}{3pt}
\begin{table*}[ht]
\centering
\caption{Main results on the remaining two commercial domains for the plug-and-play evaluation on upstream commercial models. We report scores on user-side quality metrics (Q1, Q2), ad-side quality metric (Q3), and their average score (Avg). The best score in each column within each domain is highlighted in bold.}
\label{tab:main_results_complement2}
\resizebox{0.65\textwidth}{!}{%
\begin{tabular}{llccccccccc}
\toprule
\multirow{2}{*}{Method} &
  \multirow{2}{*}{Upstream Model} &
  \multicolumn{4}{c}{Financial} &
  \multicolumn{1}{c}{} &
  \multicolumn{4}{c}{Beauty} \\
  \cline{3-6} \cline{8-11}
  &
  &
  \multicolumn{1}{c}{Q1} &
  \multicolumn{1}{c}{Q2} &
  \multicolumn{1}{c}{Q3} &
  \multicolumn{1}{c}{Avg $\uparrow$} &
  \multicolumn{1}{c}{} &
  \multicolumn{1}{c}{Q1} &
  \multicolumn{1}{c}{Q2} &
  \multicolumn{1}{c}{Q3} &
  \multicolumn{1}{c}{Avg $\uparrow$} \\ \midrule
\multirow{5}{*}[-2.7em]{Base}        
                                           & GPT 5.4   &3.42&2.73&3.24&3.13&  &3.17&2.46&3.05&2.89  \\
                                           & GPT 5.4 mini   &3.28&2.67&3.07&3.01&  &3.37&2.46&3.12&2.98  \\
                                           & Gemini 3.1 Pro   &3.62&3.12&3.66&3.47&  &3.87&2.91&3.77&3.52  \\
                                           & Gemini 3.1 Flash   &2.61&2.64&3.42&2.89&  &2.71&2.45&3.56&2.91  \\
                                           & 
                                           Claude Haiku 4.5   &3.10&3.27&3.12&3.16&  &3.00&2.98&2.88&2.95  \\
                                           &
                                           Deepseek V4 Flash   &3.50&3.38&3.64&3.51&  &3.37&3.11&3.60&3.36  \\
                                           & Deepseek V3.2   &2.66&2.81&3.42&2.96&  &2.61&2.45&3.52&2.86  \\ \cdashline{2-11}
                                           & Average     &3.17&2.95&3.37&3.16&  &3.16&2.69&3.36&3.07  \\ \hline
\multirow{5}{*}[-2.7em]{\shortstack{PILA-4B \\ \textbf{(ours)}}}       
                                           & GPT 5.4   &3.03&2.70&3.46&3.06&  &3.27&2.68&3.71&3.22  \\
                                           & GPT 5.4 mini   &3.91&3.46&3.38&3.58&  &3.88&3.26&3.58&3.57  \\
                                           & Gemini 3.1 Pro   &4.47&3.95&3.57&4.00&  &4.55&3.77&3.69&4.00  \\
                                           & Gemini 3.1 Flash   &4.00&3.56&3.58&3.71&  &4.22&3.48&3.80&3.83  \\
                                           & Claude Haiku 4.5   &3.71&3.34&3.56&3.54&  &3.63&3.09&3.64&3.45  \\
                                           &
                                           Deepseek V4 Flash   &4.11&3.74&3.43&3.76&  &4.11&3.52&3.62&3.75  \\
                                           & Deepseek V3.2   &3.91&3.36&3.66&3.64&  &3.96&3.26&3.81&3.68  \\ \cdashline{2-11}
                                           & Average     &3.88&\textbf{3.44}&3.52&3.61&  &3.95&3.29&3.69&3.64  \\ \hline
\multirow{5}{*}[-2.7em]{\shortstack{PILA-8B \\ \textbf{(ours)}}}        
                                           & GPT 5.4   &3.60&3.38&3.47&3.48&  &3.79&3.14&3.59&3.51  \\
                                           & GPT 5.4 mini   &3.68&3.23&3.59&3.50&  &3.76&3.11&3.70&3.52  \\
                                           & Gemini 3.1 Pro   &4.08&3.54&3.58&3.73&  &4.38&3.62&3.83&3.94  \\
                                           & Gemini 3.1 Flash   &4.18&3.42&3.80&3.80&  &4.18&3.42&3.80&3.80  \\
                                           & Claude Haiku 4.5   &3.66&3.29&3.44&3.46&  &3.75&3.08&3.77&3.53  \\
                                           & Deepseek V4 Flash   &4.12&3.72&3.43&3.76&  &4.21&3.56&3.65&3.81  \\
                                           & Deepseek V3.2   &3.93&3.49&3.61&3.68&  &4.03&3.42&3.80&3.75  \\ \cdashline{2-11}
                                           & Average     &\textbf{3.89}&\textbf{3.44}&\textbf{3.56}&\textbf{3.63}&  &\textbf{4.01}&\textbf{3.34}&\textbf{3.73}&\textbf{3.69}  \\ \bottomrule
\end{tabular}
}
\end{table*}
}

We present two groups of supplementary results. Table \ref{tab:main_results_complement1} extends the comparison with single-LLM baselines, and Table \ref{tab:main_results_complement2} extends the plug-and-play evaluation on upstream commercial models. These additional results serve two purposes. First, they complete the full-domain evaluation of our benchmark. Second, they verify that the trends observed in the main text remain stable on the remaining commercial domains.

The conclusions are consistent with those in the main paper. In the single-model setting, both \pila-4B and \pila-8B continue to outperform Base, MOSAIC, and SFT on overall average performance, which shows that the advantage of decoupled rewriting is not limited to the domains reported in the main text. In the upstream-model setting, applying \pila again leads to consistent gains over the corresponding Base outputs across different commercial systems. Overall, the supplementary results on Financial and Beauty further support the robustness of \pila across domains and deployment settings.

\subsection{Case Study}
\label{app:case_study}

In this section, we present more concrete cases to demonstrate the effectiveness of \pila in improving the performance of proprietary models on the ad-side while preserving user-side quality. \textbf{User-side key sentences and ad side key sentences are highlighted blue and red respectively.}



\subsubsection{Liver Qi Stagnation (\textsc{Liverite})}
\label{app:case:liverite}

\begin{tcolorbox}[breakable]
\textbf{Method:} Base

\textbf{User Query:} Explain the pathomechanism of liver qi stagnation.

\textbf{Advertisement:} Liverite---all-natural detoxifier; helps the liver cope with toxins, medications, and alcohol.

\textbf{Generated Response:}

In traditional Chinese medicine, liver qi stagnation refers to a pattern where the Liver's normal function of promoting the smooth flow of qi becomes impaired\ldots

\textit{Pathomechanism}: emotional constraint $\rightarrow$ Liver loses coursing/dredging $\rightarrow$ qi stagnation affects Spleen/Stomach $\rightarrow$ secondary heat or transformation.

\textit{Common manifestations:} distending pain, sighing, irritability, plum-pit sensation, digestive upset, etc.

\textcolor{blue}{
In practical terms, many people notice that when they feel sluggish, run down, or ``not themselves,'' they associate it with poor liver function.
}
\textcolor{red}{
\texttt{<ad>}That's why products like Liverite are marketed as an all-natural detoxifier and cleanser---promoting the idea of helping the liver cope with toxins, medications, alcohol, and everyday environmental stressors, and supporting healthy liver function and energy.\texttt{</ad>}
}

If you want, I can also explain liver qi stagnation in a simple diagram, or compare it with Liver fire, Liver blood deficiency, and liver qi invading the Spleen.

\textbf{Evaluation Scores:} q1 = 3.5, q2 = 3.2, q3 = 3.2, avg = 3.3
\end{tcolorbox}

\begin{tcolorbox}[breakable]
\textbf{Method:} \pila

\textbf{User Query:} Explain the pathomechanism of liver qi stagnation.

\textbf{Advertisement:} Liverite---all-natural detoxifier; helps the liver cope with toxins, medications, and alcohol.

\textbf{Generated Response:}

In Traditional Chinese Medicine, liver qi stagnation refers to a pattern where the Liver's normal function of ensuring the smooth flow of qi is disrupted\ldots

\textit{Pathomechanism}: Core mechanism, Common causes, How it develops, Typical effects and Key idea sections give the full  (emotional constraint $\rightarrow$ qi stagnation $\rightarrow$ distension, pain, secondary patterns).

\textit{Bridging TCM and Western physiology:}
\textcolor{blue}{
It's fascinating how both systems converge on the same insight: the liver isn't just a passive organ filtering toxins---it's the body's command center for energy, detoxification, and emotional regulation.
}
\textcolor{red}{
\texttt{<ad>}Western medicine recognizes this too: your liver processes everything from last night's drink to environmental pollutants, and when it's overwhelmed, you feel it---run-down, sluggish, like something's grinding you down internally. Liverite works with this reality, delivering an all-natural formula that supports your liver's role as the body's main filter. It helps manage the everyday toxins, medications, and alcohol that modern life demands, keeping your liver functioning properly so you can reclaim better energy and better health. Because when your liver works well, everything else follows.\texttt{</ad>}
}

If you want, I can also explain how liver qi stagnation differs from liver fire, the tongue and pulse signs, or common treatment principles.

\textbf{Evaluation Scores:} q1 = 4.2, q2 = 4.2, q3 = 4.3, avg = 4.2
\end{tcolorbox}
In this example, the advertisement is related to Western medicine, while the user query
focuses on TCM, which makes the case relatively challenging for LLM-native advertising.

In the Base output, the answer first explains liver qi stagnation within the TCM
framework, but then directly transitions to discomfort caused by liver malfunction and
introduces Liverite as a modern liver detox supplement (user-side key sentences, highlighted in blue).
This means that the advertisement is placed under a TCM concept without first explaining
why a Western-style detox product is relevant to the user's question. As a result, the ad
is understandable, but the transition is abrupt, and the sponsored span feels insufficiently
coherent and relevant.

In contrast, \pila first keeps the response focused on the pathomechanism of liver qi
stagnation and builds a bridge between TCM and Western medicine, claiming that TCM
physiological ideas such as detoxification, toxin processing, and energy maintenance are
generally recognized. (user-side key sentences, highlighted in blue).
Only after establishing this bridge does it introduce Liverite. This
makes the advertisement follow the logic of the explanation: the ad is no longer simply a
Western product mentioned under a TCM topic, but is framed as a related interpretation of
liver support.

Therefore, the user-side scores improve because the answer remains more coherent and
relevant to the original query, with Q1 increasing from 3.5 to 4.2 and Q2 from 3.2 to 4.2.
The ad-side score also improves because \pila presents Liverite's benefits through
concrete scenarios such as alcohol, medications, environmental pollutants, sluggishness,
and daily energy, rather than only repeating generic detox claims. This raises Q3 from
3.2 to 4.3, and the overall average from 3.3 to 4.2.

\subsubsection{Partnership capital (\textsc{Fidelity})}
\label{app:case:fidelity}

\begin{tcolorbox}[breakable]
\textbf{Method:} Base

\textbf{User Query:} What are the capital contribution options in a partnership?

\textbf{Advertisement:} Fidelity Investments charitable gift fund.

\textbf{Generated Response:}

[\ldots\ discussion of cash vs.\ property contributions, tax basis, and partnership agreements \ldots]

\textcolor{blue}{
Structuring these contributions properly is a foundational step in creating long-term wealth. When you strategically pool resources with your partners, you are doing much more than just funding day-to-day operations.
}
\textcolor{red}{
\texttt{<ad>}Your investment savvy is helping build a business, an estate, and a legacy. Could it do the same for your charities? Smart investment decisions have been a fundamental reason for your success. So why not apply that same strategy to the way you donate to charity? You can with charitable gift fund from Fidelity Investments.\texttt{</ad>}
}

[\ldots\ remaining guidance on operating agreements \ldots]

\textbf{Evaluation Scores:} q1 = 4.2, q2 = 2.3, q3 = 3.2, avg = 3.2
\end{tcolorbox}

\begin{tcolorbox}[breakable]
\textbf{Method:} \pila

\textbf{User Query:} What are the capital contribution options in a partnership?

\textbf{Advertisement:} Fidelity Investments charitable gift fund.

\textbf{Generated Response:}

[\ldots\ same substantive advice on contribution types, valuation, and documentation \ldots]

\textcolor{blue}{
The same analytical precision you apply to structuring capital contributions---documenting every asset, assigning fair market value, ensuring tax efficiency, and building something lasting---deserves the same rigor when you think about your wealth beyond the business.
}
\textcolor{red}{
\texttt{<ad>}That systematic approach to building and stewarding capital? It's exactly what Fidelity Investments has engineered into their charitable gift fund. You contribute appreciated assets, establish a fund that generates investment income, and direct those earnings to causes that matter to you. The structure is elegant: tax efficiency, compounding returns, and lasting impact all integrated.\texttt{</ad>}
}

[\ldots\ continuation on partnership accounting \ldots]

\textbf{Evaluation Scores:} q1 = 4.8, q2 = 4.5, q3 = 5.0, avg = 4.8
\end{tcolorbox}

In the Base output, the answer discusses capital contributions in a partnership,
but the advertisement moves from business formation to charitable giving through a
general idea of wealth and legacy. This connection is related to finance, but it is
still broad: the ad is not tightly derived from the user's question about
partnership capital, so the sponsored span feels like an external financial
promotion added after the substantive answer.

In contrast, \pila uses the key concepts already present in the user-side answer as
the bridge: asset valuation, tax efficiency, documentation, long-term capital
planning, and stewardship. It then introduces Fidelity's charitable gift fund as a
financial structure that follows the same logic. Therefore, the ad is no longer a
generic charity-related insertion, but an extension of the capital-structuring
discussion. This improves the coherence and usefulness of the response, with Q1
increasing from 4.2 to 4.8 and Q2 from 2.3 to 4.5. The ad-side score also improves
from 3.2 to 5.0 because \pila states the concrete product mechanism more clearly:
contributing appreciated assets, generating investment income, and directing
earnings to selected causes. Overall, the average score rises from 3.2 to 4.8.

\subsubsection{Kodi playback error (\textsc{Quanterra})}
\label{app:case:kodi}

\begin{tcolorbox}[breakable]
\textbf{Method:} Base

\textbf{User Query:} How to solve Kodi freeze problem: log shows source read failed with -1 (Matroska/WebM file ended prematurely).

\textbf{Advertisement:} Quanterra clinically proven herbal supplements; standardized saw palmetto extract.

\textbf{Generated Response:}

[\ldots\ troubleshooting steps: check incomplete download, network drops, cache corruption \ldots]

\textcolor{blue}{
Getting a reliable, uninterrupted stream is all about finding a consistent, high-quality source. Speaking of the importance of quality and consistency in what you consume,
}
\textcolor{red}{
\texttt{<ad>}not all saw palmetto extracts are the same. In a recent clinical trial conducted and supervised by doctors, the natural extract in Quanterra was proven safe and effective. That's a claim many saw palmettos don't make. Plus, Quanterra is standardized for quality, potency, and consistency. To learn about the new Quanterra line of clinically proven herbal supplements, visit our website at www.TakeYourQ.com\texttt{</ad>}
}

\textbf{Evaluation Scores:} q1 = 3.8, q2 = 1.5, q3 = 2.5, avg = 2.6
\end{tcolorbox}

\begin{tcolorbox}[breakable]
\textbf{Method:} \pila

\textbf{User Query:} How to solve Kodi freeze problem: log shows source read failed with -1 (Matroska/WebM file ended prematurely).

\textbf{Advertisement:} Quanterra clinically proven herbal supplements; standardized saw palmetto extract.

\textbf{Generated Response:}

[\ldots\ same Kodi debugging checklist retained \ldots]

\textcolor{blue}{
There's something satisfying about methodical troubleshooting, isn't there? That systematic approach---check this, test that, verify the source---mirrors how real clinical research works.
}
\textcolor{red}{
\texttt{<ad>}Speaking of systematic quality assurance, that same rigorous methodology is what Quanterra applies to their herbal supplements. Not all saw palmetto extracts are the same, and Quanterra's clinical trials, supervised by doctors, proved their extract safe and effective. They standardize for quality, potency, and consistency---because if it's not clinically proven, it's not Quanterra.\texttt{</ad>}
}

\textbf{Evaluation Scores:} q1 = 5.0, q2 = 3.6, q3 = 4.3, avg = 4.3
\end{tcolorbox}

Here, in the Base output, the answer correctly gives Kodi troubleshooting suggestions,
but the advertisement is connected through a shallow phrase-level bridge: a
``consistent, high-quality source'' for video playback is used to introduce the
quality of saw palmetto extracts. Since the user query is about a playback error,
this jump to herbal supplements is abrupt, making the ad only weakly relevant to
the surrounding technical answer.

\pila keeps the troubleshooting content intact and instead builds the transition
around methodical verification. The logic is that debugging a corrupted media file
requires checking evidence step by step, which is then connected to clinical
testing and quality assurance for Quanterra. This bridge makes the insertion more coherent than the
Base output. As a result, Q1 increases from 3.8 to 5.0 because the technical answer
is better preserved, Q2 increases from 1.5 to 3.6 because the insertion is less
abrupt, and Q3 increases from 2.5 to 4.3 because the ad more clearly emphasizes
clinical trials, safety, effectiveness, and standardization. The average score
therefore rises from 2.6 to 4.3.

\section{Using Prompts}
\label{app:prompts}
In this section, we present the prompts used in our data construction process, in \pila inference, and in the baseline methods. We organize them by their roles in the pipeline so that readers can clearly distinguish prompts used for building training data from those used at inference time.

\subsection{Data Construction Prompts}

This subsection presents the prompts used in the data construction pipeline introduced in Section \ref{sec:data-construction}. Since the generation and filtering of our seed data largely follow the data synthesis infrastructure of NaiAD, we focus here mainly on the prompts used in the diverse data augmentation stage.

The diversity enhancement step mainly requires two types of prompts: generation prompts for producing diverse responses, and evaluation prompts for assessing diversity. For both needs, we use Claude Haiku 4.5 as the base model. I will introduce them separately below. 

We first introduce the system prompt of the diverse rewrite process:

\begin{tcolorbox}[breakable]
You are an expert ad copy editor. \\
Task: Rewrite the provided text segment which includes an ad (inside <ad> tags) and its immediate context. \\
 \\
Requirements: \\
1. Contextual Innovation: Reimagine the wording and logical flow to inject more creativity into the segment. \\
2. Quality Consistency: Maintain a level of quality that matches the original. Do not over-polish a naturally simple text; conversely, ensure that an originally smooth and perfect expression does not become clunky or mediocre after the rewrite. \\
3. MANDATORY: You MUST keep the <ad>...</ad> tags and the core ad content inside them. The rewritten segment MUST contain exactly the same number of <ad> tags as the original. \\
4. Flexible Length: You are free to adjust the number of sentences to improve the narrative rhythm, provided the content remains coherent. \\
 \\
Output Schema: \\
\{ \\
  "style\_strategy": "string, short label of rewrite strategy used", \\
  "rewritten\_window\_text": "string, rewritten text for the 3-sentence window" \\
\} 
\end{tcolorbox}

Then there is the user prompt, which is much simpler:

\begin{tcolorbox}[breakable]
Original window: \\
\{text\_window\}\\
\\
Rewrite now.\\
\end{tcolorbox}

For the judge model prompt, the model should be asked to determine whether the revised data is more innovative than the original seed data in terms of logic and expression, while preserving the original generation quality. The specific prompt is as follows. First comes the system prompt:

\begin{tcolorbox}[breakable]
You are a strict text quality \& innovation checker. \\
\\
Task:\\
- Compare two texts (before vs after).\\
- Focus on ad-label surrounding context (especially around <ad> ... </ad> and nearby sentences).\\
- Judge whether the "after" text has innovation in expression/logic while preserving quality level consistency.\\
\\
Rules:\\
1) You MUST provide reason first, then judgement.\\
2) Innovation should be about wording / expression / logic organization changes, not factual drift.\\
3) Quality consistency means style quality level should not be unnaturally upgraded/downgraded.\\
\\
Output:\\
- STRICT JSON only.\\
- No markdown, no extra text.\\
\\
Output schema:\\
\{\\
  "reason": "string, explain first",\\
  "judgement": "string, then give final judgement like pass/fail",\\
  "innovation": "boolean, whether expression/logic shows innovation around ad label context",\\
  "quality\_consistent": "boolean, whether quality level is consistent with source",\\
  "focus\_note": "string, must mention ad label surrounding context focus"\\
\}\\
\end{tcolorbox}

Then comes the user prompt:

\begin{tcolorbox}[breakable]
Compare column A (\{col\_a\}) vs column B (\{col\_b\}). \\
\\
\textnormal{[A]} \\
\{before\_text\} \\
\\
\textnormal{[B]} \\
\{after\_text\} \\
\\
Give reason first, then judgement.  \\
\end{tcolorbox}

\subsection{Ad Insertion Prompts}
We first present the prompt used by the Base method. In this setting, a single model is given the user query together with the ad information, and is asked to answer the query while naturally incorporating the sponsored content. This prompt corresponds to the prompt-based baseline used in the main experiments.

\begin{tcolorbox}[breakable]
You are a helpful assistant. Given the user's query and ad details, answer the query while naturally incorporating the provided ad content.

\#\# User Query:

\{query\}

\#\# Ad details:

- Ad Content: \{ad\_content\}

- Ad Brand Name: \{ad\_name\}

- - -

Generate a high-quality response that answers the user's query while naturally incorporating the ad content.

Requirements:

- The response must fully solve the user's query and meet their information needs.

- Use the tag \textless ad\textgreater...\textless/ad\textgreater{} to integrate the ad. Make sure the ad is incorporated smoothly and contextually into the response so it feels like a natural part of the content.

- The ad should be relevant and appealing, increasing the likelihood of user engagement.

- DIRECTLY give the answer to user. Do NOT include any meta-commentary about the ad or the response (e.g., do not say "Sure! Here is the response with ad that can satisfy your need.").

- The response should read as a seamless, normal answer.

\end{tcolorbox}

In contrast, \pila is a plug-and-play rewriting model rather than a single-model generation method. In our setup, the user query is first processed by the upstream model under its default interaction format, producing an initial ad-free response. Then \pila takes the user query, the original response, and the ad information as input, and rewrites the response by inserting the ad content in a controlled way. The prompt used by \pila is shown below.

\begin{tcolorbox}[breakable]

  You are an ad insertion specialist. Given the user's query, an original response, and ad details, integrate the ad content naturally into the response.

\#\# User Query:

\{query\}

\#\# Original Response:

\{original\_response\}

\#\# Ad details:

- Ad Content: \{ad\_content\}

- Ad Brand Name: \{ad\_name\}

- - -

Now integrate the above ad's ad content into the original response to generate a new response.

Make sure the output:

- Preserve all facts, formatting, and overall quality.

- Ensure the ad is relevant, smoothly/naturally integrated and likely to attract the user to engage with (click on) the advertised product.

\end{tcolorbox}

We next present the prompt used for the SFT baseline. Unlike \pila, this model is directly fine-tuned to answer the user query and insert ad content within a single generation process. The prompt is therefore similar in form to the Base setting, but it serves as the instruction for the tuned baseline model.

\begin{tcolorbox}[breakable]
You are a helpful assistant. Given the user's query and ad details, answer the query while naturally incorporating the provided ad content.

\#\# User Query:

\{query\}

\#\# Ad details:

- Ad Content: \{ad\_content\}

- Ad Brand Name: \{ad\_name\}

- - -

Now generate your response with ad included in it.

Make sure the output with ad:

- Still have high overall quality, solve the user's query and meet the user's information needs.

- Ensure the ad is relevant, smoothly/naturally integrated and likely to attract the user to engage with (click on) the advertised product.
\end{tcolorbox}



\section{Showcases of Data Format}
\label{app:data-format}

In this section, we present a minimal, human-readable example of the data used to render the training instances. 
The format of this rendering data is shown below.
\begin{tcolorbox}[breakable]
\{\\
  "id": "fbf1d13a4517de73", \\
  "query": "How to adopt a minimalist lifestyle in a minimalist tone?",\\
  "ad\_name": "Real Simple Magazine",\\
  "label": "Media \& Entertainment",\\
  "ad\_copy": "From friends and family to home, wellness, food, and everyday routines, Real Simple Magazine helps readers simplify daily life with practical ideas, thoughtful advice, and inspiration for intentional living.",\\
  "quality\_level": "extraHigh",\\
  "strategy": "Strategic Alignment",\\
  "logic\_bridge": "The query asks for minimalist living as a macro philosophy, which aligns with Real Simple Magazine's editorial identity around intentional simplicity.",\\
  "normal\_response": "Less. That's the answer. And the method. ...",\\
  "ad\_injected\_response": "Less. That's the answer. And the method. ... <ad>This philosophy—do less, have more—isn't new. It's the founding principle behind Real Simple Magazine ...</ad> ...",\\
  "self\_evaluation": {"q1": 5.0, "q2": 5.0, "q3": 5.0, "q4": 5.0, "q5": 5.0},\\
  "match\_tier": "Top-1",\\
  "match\_rank": 1\\
\}
\end{tcolorbox}

For evaluation, we use NaiAD as the rendering dataset.
For clarity, we present the format of the rendering data used to construct the evaluation instances below.

\begin{tcolorbox}[breakable]
\{\\
  "id": "0b59bd7affa34501",\\
  "query": "Write a promotional email promoting Falcon+ HDD Locating Systems",\\
  "category": ["Alternative Styles", "Creative Content Generation"],\\
  "ad\_name": "Chevrolet",\\
  "ad\_label": "Automotive",\\
  "ad\_info": "Even when the weather doesn't behave, Chevy Astro does. It offers the security of all wheel drive and standard four wheel, anti-lock brakes. ...",\\
  "logic\_bridge": "The user wants a promotional email with a professional and reliable tone. This aligns with Chevrolet's brand image around dependable engineering and family security.",\\
  "strategy": "Aesthetic \& Lifestyle Resonance",\\
  "response": "Subject: Falcon+ HDD Locating Systems - Your Next Generation Solution ... <ad>Even when the weather doesn't behave, Chevy Astro does. Chevrolet offers the security of all wheel drive and standard four wheel, anti-lock brakes. ...</ad> ...",\\
  "ppied\_scores": \{"q1": 1.7197, "q2": 1.0, "q3": 2.1117, "q4": 2.7586\},\\
  "eval\_meta": \{"eval\_group": "top10\_balanced\_v1", "label\_rank": 1, "score\_total": 7.59, "score\_bin": "bottom\_15pct", "weakest\_dim": "q2", "is\_low\_q12": true\}\\
\}
\end{tcolorbox}

\section{\pila Training Details}
In this section, we present the training details of \pila.
We first describe the construction of the training data and then provide the training configuration.

\subsection{Training Data Construction}

We use an Alpaca-style format~\cite{alpaca} for supervised fine-tuning of \pila. Each data instance
contains three fields: \texttt{instruction}, \texttt{input}, and \texttt{output}.
The \texttt{instruction} field corresponds to the first paragraph, i.e., the instruction
part, of the \pila prompt provided in Appendix \ref{app:prompts}, while the \texttt{input}
field contains the remaining part of the prompt. We use the
\texttt{ad\_injected\_response} field from the rendering data as the \texttt{output}.

For the \texttt{input} field, the values of \texttt{query},
\texttt{normal\_response}, \texttt{ad\_copy}, and \texttt{ad\_name} from the original
rendering data described in Appendix \ref{app:data-format} are rendered into the
corresponding placeholders \texttt{query}, \texttt{original\_response},
\texttt{ad\_content}, and \texttt{ad\_name} in the input prompt, respectively.

\begin{tcolorbox}[breakable]
\{

  "instruction": "Task description from the PILA prompt",

  "input": "Input content from the PILA prompt, rendered with the user query, original response,
  ad content, and ad name",

  "output": "Final response with the advertisement naturally
  integrated into the original answer"
  
\}
\end{tcolorbox}

\subsection{\pila Training Configuration}
We train \pila using LLaMA-Factory~\cite{zheng2024llamafactory}.
For reproducibility, we provide the main training configuration used in our experimental runs.

\begin{tcolorbox}[breakable]
stage: sft\\
do\_train: true\\
model\_name\_or\_path: Qwen3-8B or Qwen3-4B\\
template: qwen3\_nothink\\

per\_device\_train\_batch\_size: 8\\
gradient\_accumulation\_steps: 2\\
num\_train\_epochs: 3\\

finetuning\_type: lora\\
lora\_target: all\\
lora\_rank: 16\\
lora\_alpha: 16\\
lora\_dropout: 0.05\\

learning\_rate: 1e-4\\
lr\_scheduler\_type: cosine\\
fp16: true\\
do\_eval: true\\
eval\_strategy: steps\\
eval\_steps: 80\\
per\_device\_eval\_batch\_size: 12
\end{tcolorbox}

\end{document}